\documentclass[letterpaper]{article} 
\usepackage{aaai2026}  
\usepackage{times}  
\usepackage{helvet}  
\usepackage{courier}  
\usepackage[hyphens]{url}  
\usepackage{graphicx} 
\usepackage{natbib}  
\usepackage{caption} 
\usepackage{algorithm}
\usepackage{algorithmic}
\usepackage{booktabs}
\usepackage{amsmath}
\usepackage{amssymb}
\usepackage{subcaption}

\usepackage{newfloat}
\usepackage{listings}
\DeclareCaptionStyle{ruled}{labelfont=normalfont,labelsep=colon,strut=off} 
\floatstyle{ruled}
\newfloat{listing}{tb}{lst}{}
\floatname{listing}{Listing}
\title{AVM: Towards Structure-Preserving Neural Response Modeling in the Visual Cortex Across Stimuli and Individuals}
\author{
    Qi Xu\textsuperscript{\rm 1}, 
    Shuai Gong\textsuperscript{\rm 1},
    Xuming Ran\textsuperscript{\rm 2}\thanks{Corresponding author:Xuming Ran},
    Haihua Luo\textsuperscript{\rm 1,3},
    Yangfan Hu\textsuperscript{\rm 4}\\
}
\affiliations{
    \textsuperscript{\rm 1}School of Computer Science and Technology, Dalian University
    of Technology,\\
    \textsuperscript{\rm 2}National University of Singapore,\\
    \textsuperscript{\rm 3}Faculty of Information Technology, University of Jyväskylä,\\
    \textsuperscript{\rm 4}School of Information Technology and Artificial Intelligence, Zhejiang University of Finance and Economics,\\
    ranxuming@gmail.com\\
    
}

\usepackage{bibentry}

\begin{document}

\maketitle

\begin{abstract}
While deep learning models have shown strong performance in simulating neural responses, they often fail to clearly separate stable visual encoding from condition-specific adaptation, which limits their ability to generalize across stimuli and individuals. We introduce the \textbf{A}daptive \textbf{V}isual \textbf{M}odel (\textbf{AVM}), a structure-preserving framework that enables condition-aware adaptation through modular subnetworks, without modifying the core representation. AVM keeps a Vision Transformer-based encoder frozen to capture consistent visual features, while independently trained modulation paths account for neural response variations driven by stimulus content and subject identity. We evaluate AVM in three experimental settings, including stimulus-level variation, cross-subject generalization, and cross-dataset adaptation, all of which involve structured changes in inputs and individuals. Across two large-scale mouse V1 datasets, AVM outperforms the state-of-the-art V1T model by approximately 2\% in predictive correlation, demonstrating robust generalization, interpretable condition-wise modulation, and high architectural efficiency. Specifically, AVM achieves a 9.1\% improvement in explained variance (FEVE) under the cross-dataset adaptation setting. These results suggest that AVM provides a unified framework for adaptive neural modeling across biological and experimental conditions, offering a scalable solution under structural constraints. Its design may inform future approaches to cortical modeling in both neuroscience and biologically inspired AI systems.
\end{abstract}


\section{Introduction}
Understanding the computational mechanisms of neurons in the visual system—particularly how they respond to natural image stimuli—remains a central challenge in sensory neuroscience \cite{carandini2005we,yamins2013hierarchical,mcintosh2016deep,yang2020artificial,zhang2025toward2}. Modeling the response patterns of the primary visual cortex (V1) has become an effective strategy, bridging biological vision with machine perception \cite{klindt2017neural,sinz2019engineering,lurz2020generalization,li2023v1t}.

Recent advances in deep learning have significantly improved our ability to predict V1 responses. Traditional methods, including generalized linear models and shallow multi-layer networks, have been surpassed by CNNs, which extract nonlinear stimulus features with greater fidelity \cite{yamins2014performance,cadena2019deep}. More recently, Vision Transformer (ViT) models have shown superior representational power, with architectures like V1T \cite{li2023v1t} setting new benchmarks for mouse V1 modeling by coupling a ViT backbone with behavioral feature modules.

However, these models tend to conflate stable representation learning with response adaptation, making them brittle under condition changes such as stimulus shifts, individual variability, or environmental perturbations. Once trained, such monolithic models require full retraining to adjust to new conditions, limiting their generalization and interpretability. In contrast, biological visual systems exhibit a compelling duality: they preserve a stable structural organization while flexibly modulating responses based on context, internal state, or inter-subject variability \cite{franke2022state,cheng2022vision}. This motivates a critical modeling challenge: \textbf{how to reconcile structural stability with functional flexibility} in neural response prediction.

To address this, we propose the Adaptive Visual Model (AVM)—a framework grounded in the principle of \textbf{structure-function decoupling}. Unlike prior approaches that entangle representation and modulation, AVM introduces an explicit architectural separation: a shared, frozen encoder captures invariant structural representations, while lightweight, condition-aware modules enable flexible modulation of neural responses. This design not only reflects biological organization but also supports \textbf{scalable and efficient adaptation} to diverse conditions—input shifts, individual differences, and environmental changes—without altering the core encoding structure.
We evaluate AVM on two large-scale mouse V1 datasets and test its generalization ability across three major challenges in cortical modeling: stimulus changes, inter-individual differences, and cross-dataset shifts. In all scenarios, AVM consistently delivers accurate and interpretable neural predictions, achieving high performance through localized adaptation while maintaining a stable representational backbone. These results suggest that AVM provides a robust and biologically grounded solution for condition-aware neural response modeling.Our contributions are threefold:

\begin{itemize}

\item We propose AVM, a condition-aware cortical modeling framework grounded in structure-function decoupling, which separates stable representation encoding from dynamic response modulation;

\item AVM introduces modular control subnetworks that enable localized response adaptation under frozen representational backbones, satisfying biological constraints of structural stability and contextual flexibility;

\item AVM demonstrates scalable and interpretable generalization across input domains, individual anatomy, and environmental conditions, without requiring full model retraining.

\end{itemize}

\section{Related Work}

The modeling of neural responses in the visual cortex dates back to Hubel and Wiesel’s discovery of simple and complex cells \cite{hubel1962receptive, carandini1999linearity, carandini2005we, batty2017multilayer, ponce2019evolving, billeh2020systematic, bashivan2019neural, bao2020map, zhang2021rectified}. Early models were linear, describing neuronal tuning to visual features like orientation and contrast \cite{jones1987two, olshausen1996emergence}, later extended by nonlinear models such as the energy model \cite{adelson1985spatiotemporal}, LN model \cite{jones1987two}, and LN-LN cascades \cite{heeger1992half}.

Traditional models, while useful, struggle to generalize to complex stimuli. With increased computational power and large datasets, deep learning approaches, especially CNNs and ViTs, have become dominant for modeling visual cortex responses \cite{margalit2023unifying, klindt2017neural, lurz2020generalization, cotton2020factorized, ran2021deep, franke2022state, li2023v1t, du2024towards, deng2024predicting, zhang2025spike}. These models typically follow two paradigms: task-driven models use pretrained networks for object recognition and readout modules to predict neural responses \cite{yamins2014performance, cadieu2014deep, cadena2019deep}, while data-driven models train directly from neural recordings without supervision, learning shared representations across animals \cite{klindt2017neural, lurz2020generalization, cadena2019deep, franke2021behavioral, franke2022state, li2023v1t, zhang2025toward}.

Additionally, Franke et al. \cite{franke2022state} integrated behavioral information with visual stimuli, demonstrating improved neural response prediction by combining behavioral data with visual inputs to capture dynamic neural fluctuations.

\section{Method}

\begin{figure*}[ht]
\begin{center}
\includegraphics[width=\linewidth]{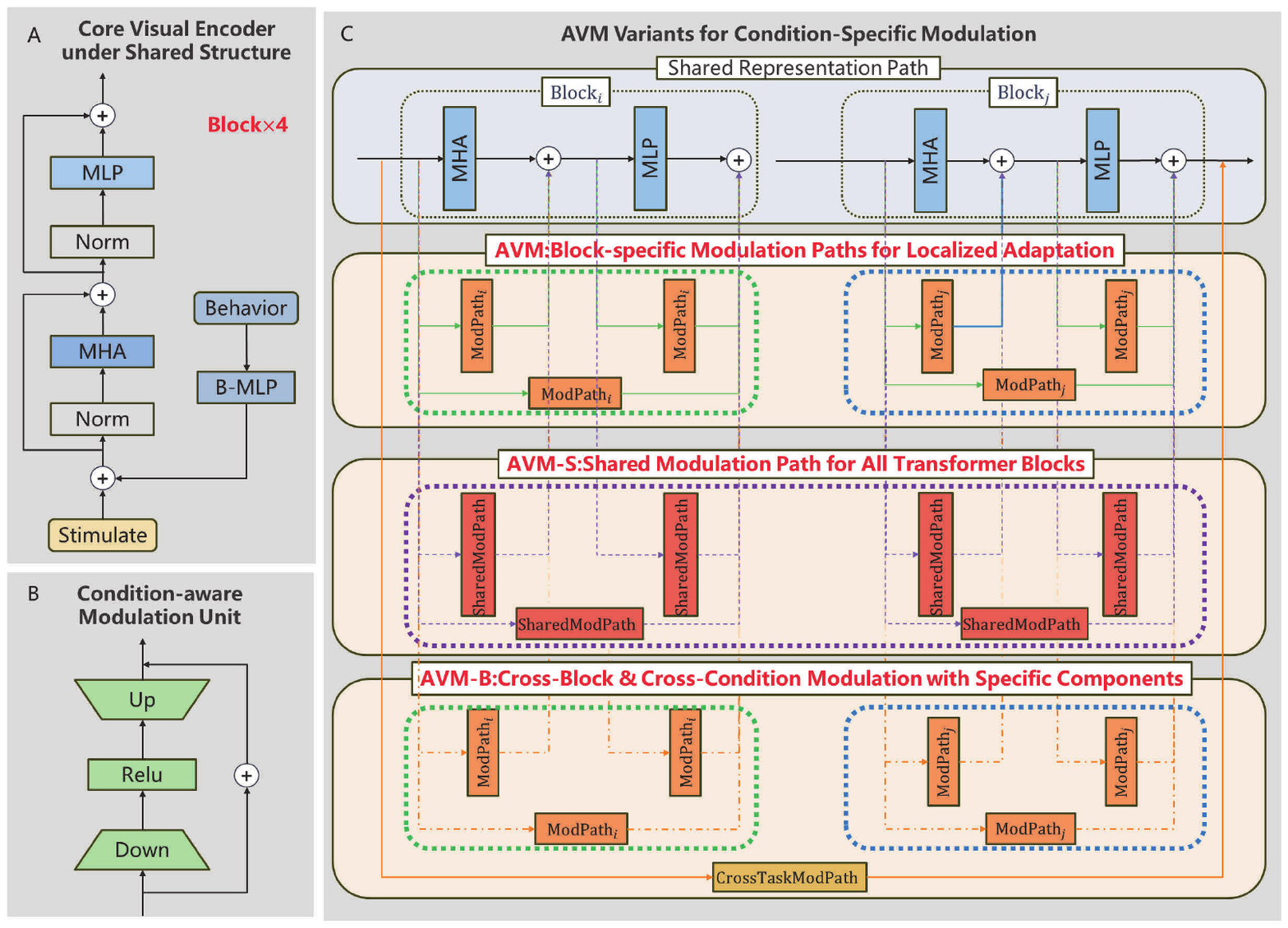}
\caption{AVM model architecture and condition-specific modulation variants. (A) The main network encodes stable visual representations under a consistent architecture. 
(B)  Condition-aware Modulation Unit(CAMU): A lightweight feedforward module with a bottleneck structure, serving as the basic modulation component across AVM variants.
(C) Condition-Specific Modulation Variants:
AVM Employs block-specific modulation paths for localized response adaptation.
AVM-S  Shares a single modulation path across all Transformer blocks, enabling parameter-efficient tuning.
AVM-B Introduces Cross-CAMU to support cross-block and cross-condition transfer, modeling higher-level adaptation interactions.
}
\label{fig:AVM}
\end{center}
\end{figure*}

\subsection{Overview: Structure-Function Decoupling in AVM}
We propose AVM (Adaptive Visual Modeling), a structure-function decoupled framework for cortical response modeling. It comprises a frozen Transformer-based visual encoder and a set of condition-aware modulation modules inserted in parallel, form ing a dual-path structure:

\begin{itemize}
\item \textbf{Shared Representation Path:} encodes invariant visual features common across subjects, stimuli, and domains;
\item \textbf{Condition-Specific Modulation Path:} applies flexible, lightweight transformations that adjust neural responses under different conditions without altering the core encoder.
\end{itemize}

This design supports interpretable, scalable adaptation while preserving representation consistency.

\subsection{Core Encoder: Stable Visual Representation}

As shown in Fig.~\ref{fig:AVM}A, the main network is a 4-layer ViT backbone (from V1T\cite{li2023v1t}), frozen throughout training. The model receives visual input $\mathbf{x}$ and behavior variables $\mathbf{b}$. The behavior signal is first embedded via a B-MLP and added to the input before processing with attention and MLP layers.

Each block contains two standard Transformer components: multi-head attention (MHA) and MLP. Their outputs form the base visual representation $\mathbf{f}$ for each location, computed as:
\begin{align}
    b &\gets b_{prev} + \operatorname{MLP}_{\text{behavior}}\left(b\right), \label{eq1} \\
    a_i &\gets \operatorname{MHA}\left(x_i + b\right) + x_i, \label{eq2} \\
    f_i &\gets \operatorname{MLP}\left(a_i\right) + a_i, \label{eq3}
\end{align}
where $\mathbf{x}_i$ is the patch embedding of the image input, $\mathbf{b}$ is the behavior vector, and $\mathbf{f}_i$ is the resulting visual representation at each layer. This frozen backbone ensures a stable and interpretable representational structure across all adaptation settings.

\subsection{Condition-Aware Modulation Unit (CAMU)}
To achieve condition-specific flexibility, AVM introduces \textbf{Condition-aware Modulation Units} (Fig.~\ref{fig:AVM}B) into each Transformer block. These lightweight modules reshape activations via bottleneck-style feedforward layers:
\begin{equation}
    \operatorname{CAMU}(x)=x+\operatorname{Up}(\operatorname{ReLU}(\operatorname{Down}(x)))
\end{equation}

Each ViT block integrates three modulation modules placed at key positions: after the attention output, the MLP output, and the block output. These modules enable localized context-sensitive adjustments under a shared structure. The revised computation becomes:

\begin{align}
    b &\gets b_{prev} + \operatorname{MLP}_{\text{behavior}}\left(b\right), \label{eq4} \\
    a_i &\gets \operatorname{MHA}\left(x_i + b\right) + x_i + \operatorname{CAMU}_{\operatorname{1}}\left(x_i\right), \label{eq5} \\
    f_i &\gets \operatorname{MLP}\left(a_i\right) + a_i + \operatorname{CAMU}_{\operatorname{2}}\left(a_i\right), \label{eq6} \\
    f &\gets f_i + \operatorname{CAMU}_{\operatorname{3}}\left(x_i\right), \label{eq7}
\end{align}

These independently trainable controllers modulate neural responses without disrupting the stable encoder stream, enabling modular, interpretable adjustments.

\subsection{Readout: Neuron-Wise Prediction}
We employ a Gaussian readout module as proposed by Lurz et al.\cite{lurz2020generalization}. Each neuron is represented by a learned 2D Gaussian position $(\mu, \sigma)$, which defines a location-sensitive sampling from the visual feature map $\mathbf{f}$. A linear layer projects sampled features to neural response space. An ELU activation with offset ensures positivity:
\begin{equation}
    \hat{y}_{n}=\operatorname{ELU}\left(\mathbf{w}_{n}^{\top} \cdot f\left(\mu_{n}\right)\right)+1
\end{equation}
 
\subsection{Modulation Variants: Adaptation Strategies}
As shown in Fig.~\ref{fig:AVM}C, AVM supports three structural variants to flexibly support various adaptation needs:

\begin{itemize}
    \item \textbf{AVM}: Each block contains a unique modulation subnetwork, enabling fine-grained, layer-wise response control for highly heterogeneous input or subject conditions.

    \begin{align}
    h_i = \text{Block}_i(h_{i-1}) + \text{ModPath}_i(h_{i-1})
    \end{align}

    \item \textbf{AVM-S}: All blocks share the same modulation subnetwork, enforcing a consistent adaptation rule across layers. This design favors parameter efficiency and coherence in response shifts.

    \begin{equation}  
    \begin{split}    
    h_i = \text{Block}_i(h_{i-1}) + \text{SharedModPath}(h_{i-1})
    \end{split}
    \end{equation}

    \item \textbf{AVM-B}: Beyond intra-layer adaptation, this variant adds inter-layer modulation to propagate context across hierarchical stages. Suitable for multi-level adaptation tasks such as cross-domain generalization.

    \begin{align}
    h_i &= \text{Block}_i(h_{i-1}) + \text{ModPath}_i(h_{i-1}) \\
    h_{ij} &= \text{Block}_j(h_i) + \text{ModPath}_j(h_i) \\
    h_{j} &= h_{ij} + \text{CrossTaskModPath}(h_{i-1})
    \end{align}
\end{itemize}
These variants enable AVM to flexibly adjust its adaptation strategy under different degrees of condition complexity while maintaining a stable representational backbone.

\begin{figure*}[h]
\begin{center}
\centerline{\includegraphics[scale=0.55]{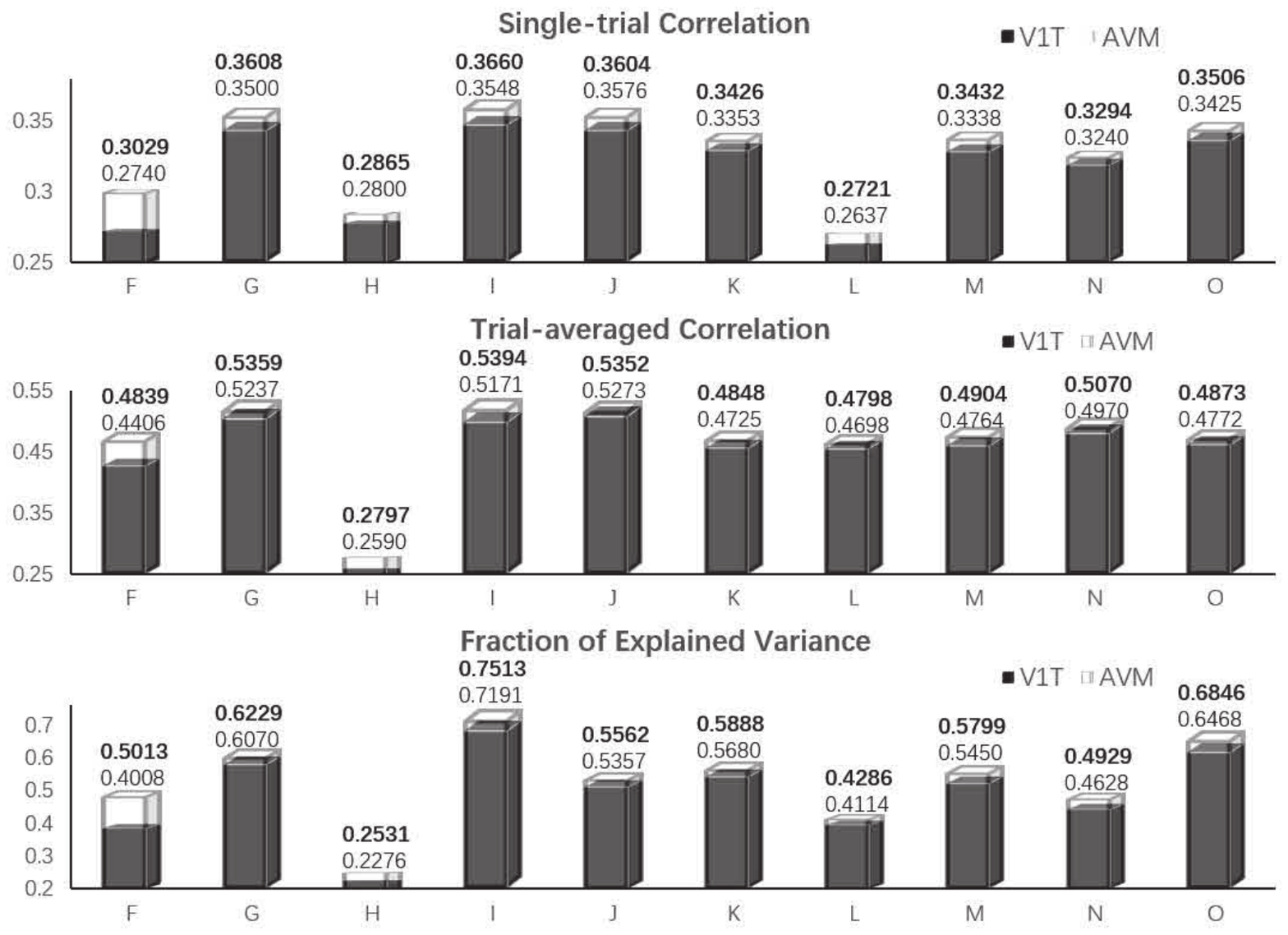}}  

\caption{AVM consistently improves individual-level neural prediction.
Evaluation results on Dataset-F for each individual mouse (F–O), comparing AVM and baseline V1T. Three metrics are reported: single-trial correlation (\textit{top}), trial-averaged correlation (\textit{middle}), and fraction of explained variance (FEVE, \textit{bottom}). AVM achieves consistent gains across all individuals.}
\label{F-single-mouse}
\end{center}
\end{figure*}

\section{Experimental Settings}

\subsection{Datasets}
This study uses two datasets for validation: the Sensorium dataset \cite{willeke2022sensorium} and the dataset of mouse primary visual cortex responses \cite{franke2022state}.

The Sensorium dataset (denoted as Dataset S) records neural activity from over 7,000 neurons in the V1 region of five mice (labeled A–E) using two-photon calcium imaging. Each mouse was presented with grayscale natural images from ImageNet ($x_{\text{image}} \in \mathbb{R}^{1 \times h \times w}$), totaling 25,100 unique images. 
Among these, 5,000 images were used for training, while 100 images were each repeated 10 times to construct the test set. For each mouse, the neural responses of $m_i$ neurons were recorded across $n$ repetitions, resulting in $m_i \times n$ total neural response samples. In addition to neuronal data, Dataset S includes anatomical coordinates for each neuron and four behavioral variables: pupil diameter, its temporal derivative, the 2D pupil center coordinates, and locomotion speed.

The Franke dataset (denoted as Dataset F) captures V1 responses from ten mice (labeled F–O) to both grayscale and color images from ImageNet. The training set consists of 4,500 unique color images ($x_{\text{image}} \in \mathbb{R}^{2 \times h \times w}$) and 750 grayscale images ($x_{\text{image}} \in \mathbb{R}^{1 \times h \times w}$), while the test set comprises 100 color and 50 grayscale images. Neural recordings were obtained from 1,000 neurons across the ten mice. For consistency with our model's input requirements, all color images from Dataset F were converted to grayscale during preprocessing.

\subsection{Training Protocol}
The AVM training process follows a two-phase strategy that reflects the model’s core design philosophy: separating stable encoding and adaptive modulation.

\textbf{Phase 1 (Backbone Pretraining)}: The main network (ViT-based backbone) and readout module are jointly trained on a designated pretraining dataset (e.g., Dataset S) using all available parameters. This phase models stable visual feature encoding shared across conditions.

\textbf{Phase 2 (Subnetwork Adaptation)}: The backbone is frozen, and a condition-specific sub-network (adapter modules) is trained on a new condition (e.g., different dataset or subject). This reflects the AVM design for localized response modulation without altering the shared representation.

We use the AdamW optimizer and the Poisson loss function:
\begin{equation}\label{eq12}
    \mathcal{L}_{m}^{\text{Poisson}}(r, o) = \sum_{t=1}^{n_t} \sum_{i=1}^{n_m} \left( o_{i,t} - r_{i,t} \log o_{i,t} \right),
\end{equation}
where $r_{i,t}$ and $o_{i,t}$ are the true and predicted responses for neuron $i$ in trial $t$.

All models are trained with a batch size of 16, an initial learning rate of 0.0016, and up to 400 epochs. We apply early stopping and learning rate decay (factor 0.3) when validation loss plateaus for 10 epochs. The same training schedule is used in both phases, except that in Phase 2 only the sub-network parameters are updated.

\subsection{Evaluation Metrics} 
The predictive performance of our model is evaluated using three metrics: single-trial correlation, average trial correlation, and the model's Fraction of Explained Variance (FEVE) \cite{willeke2022sensorium}.

Single-trial correlation considers the inter-experiment variability of the same visual stimulus presented multiple times in the test set, providing a detailed measure of the model's prediction accuracy:
\begin{equation}\label{eq13}
\rho_{trial} (r, o)=\frac{\sum_{i, j}\left(r_{i, j}-\bar{r}\right)\left(o_{i, j}-\bar{o}\right)}{\sqrt{\sum_{i, j}\left(r_{i, j}-\bar{r}\right)^{2} \sum_{i, j}\left(o_{i, j}-\bar{o}\right)^{2}}},
\end{equation}
where $r_{i,j}$ represents the true response of a single neuron to a single stimulus presentation, $\bar{r}$ is the average response across all repetitions of the image, $o_{i,j}$ is the predicted response of the neuron to the same stimulus, and $\bar{o}$ is the average predicted response across all repetitions of the image.

Average trial correlation is calculated by considering the neuron responses $r_{ij}$ for image $i$ and repetition $j$ , and their predicted values $o_i$. The correlation between the predicted responses and the average neuronal response $r_i$ for image $i$ is computed as follows:
\begin{equation}\label{eq14}
\rho_{avg} (r, o)=\frac{\sum_{i, j}\left(\bar{r_{i}}-\bar{r}\right)\left(o_{i}-\bar{o}\right)}{\sqrt{\sum_{i, j}\left(\bar{r_{i}}-\bar{r}\right)^{2} \sum_{i, j}\left(o_{i}-\bar{o}\right)^{2}}},
\end{equation}
where $\bar{r_{i}}$ denotes the average response of $r_{i,j}$ over the $j$ repeated experiments.
 
FEVE (Fraction of Explained Variance) measures the proportion of variance in the neural responses explained by the model relative to the total variance, assessing the model's ability to explain neural activity while excluding the influence of random noise or unexplained variability:
\begin{equation}\label{eq15}
\text { FEVE }=1-\frac{\frac{1}{N} \sum_{i, j}\left(r_{i,j}-o_{i}\right)^{2}-\sigma_{\varepsilon}^{2}}{\operatorname{Var}[\mathbf{r}]-\sigma_{\varepsilon}^{2}}.
\end{equation}
where $\operatorname{Var}[\mathbf{r}]$ is the total response variance computed across all $N$ trials and $\sigma^{2}_{\epsilon} = E_{i}[Var_{j}[r|x]]$ is the observation noise variance computed as the average variance across responses to repeated presentations of the stimulus $x$.

\section{Results}
\subsection{Condition-Driven Local Response Adjustment}



\begin{figure*}[h]
\centering
\begin{subfigure}[t]{0.32\textwidth}
    \centering
    \includegraphics[width=\linewidth]{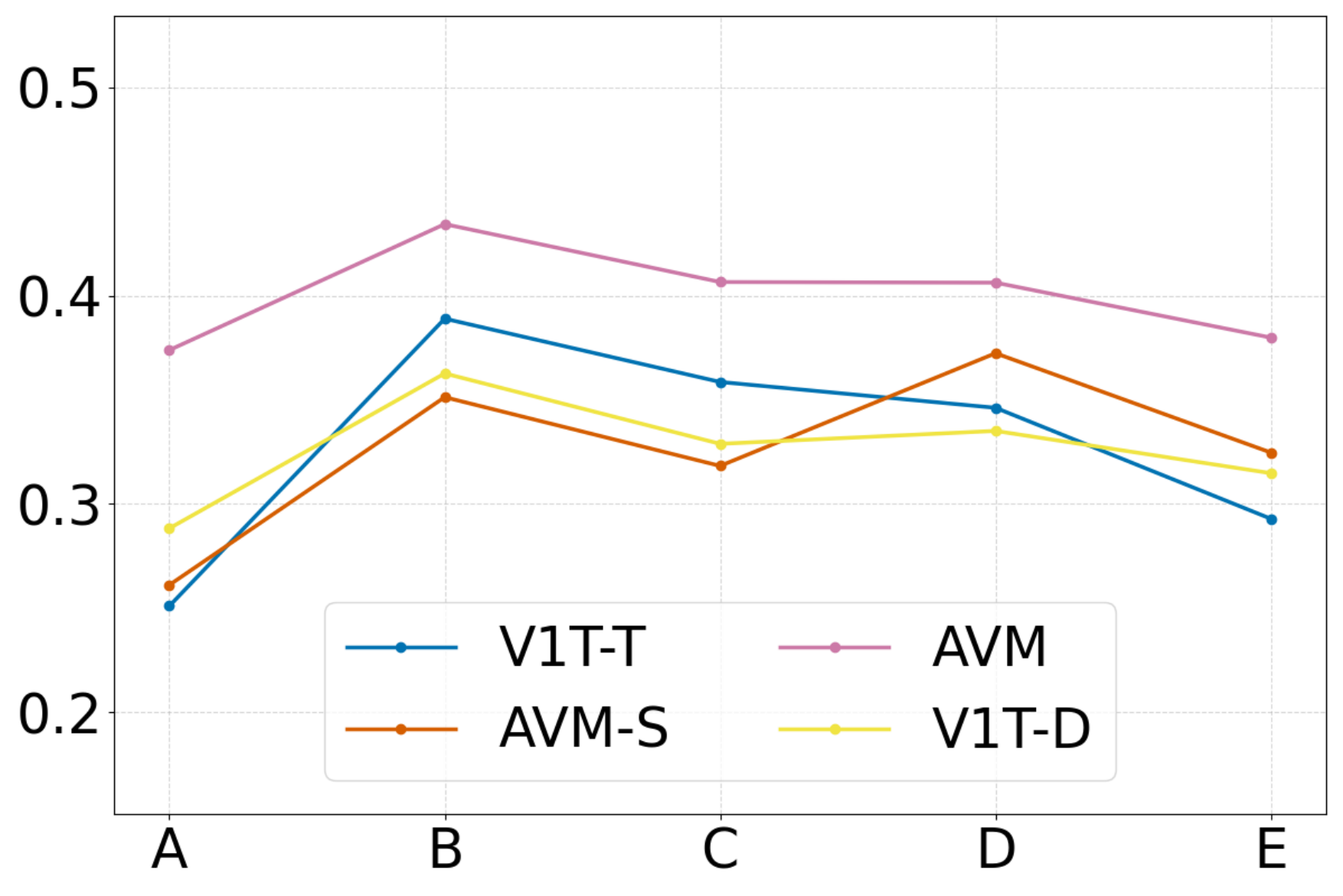}
    \vspace{0.5em} 
    \includegraphics[width=\linewidth]{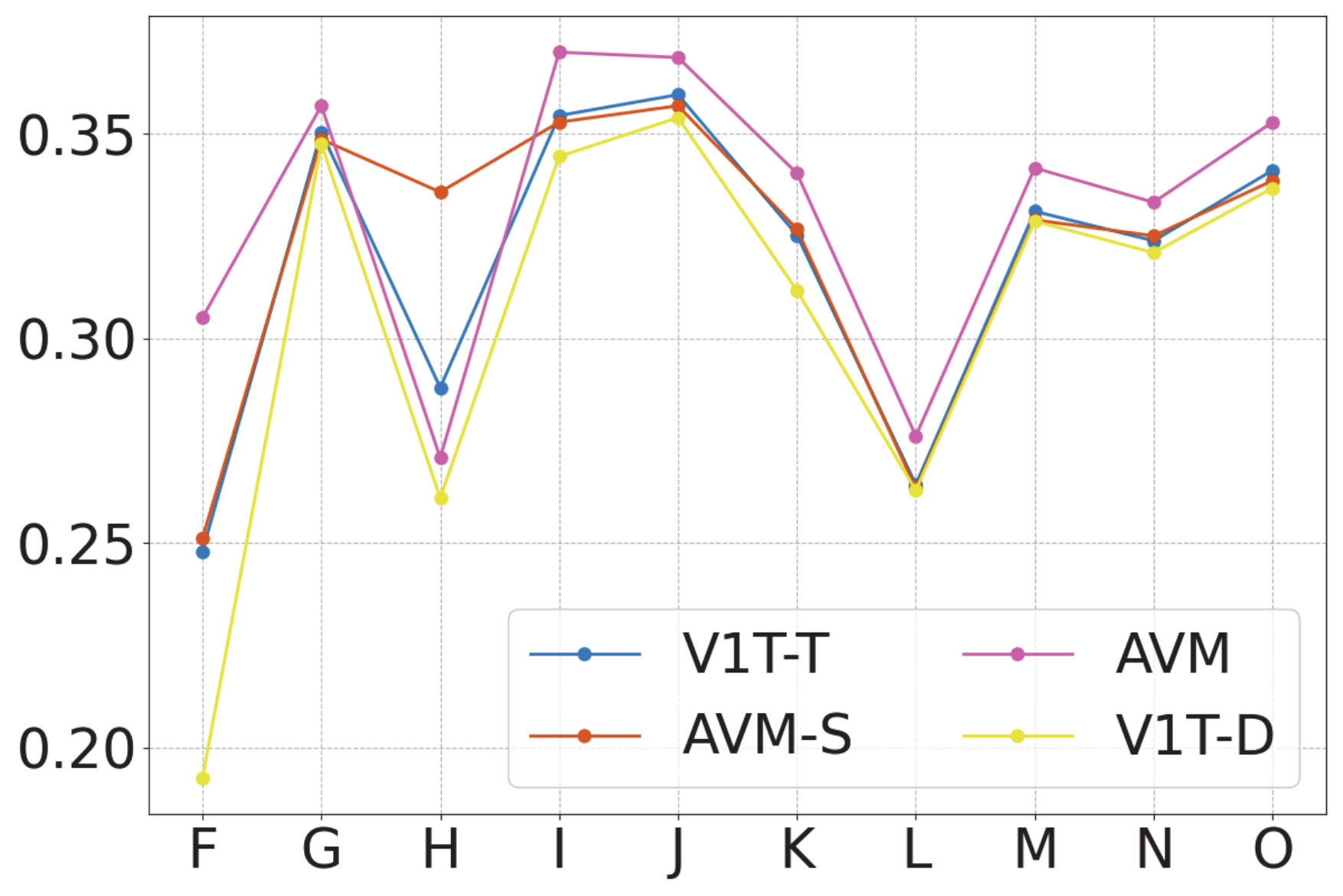}
    \caption*{(a) Single-Trial Correlation}
\end{subfigure}
\hfill
\begin{subfigure}[t]{0.32\textwidth}
    \centering
    \includegraphics[width=\linewidth]{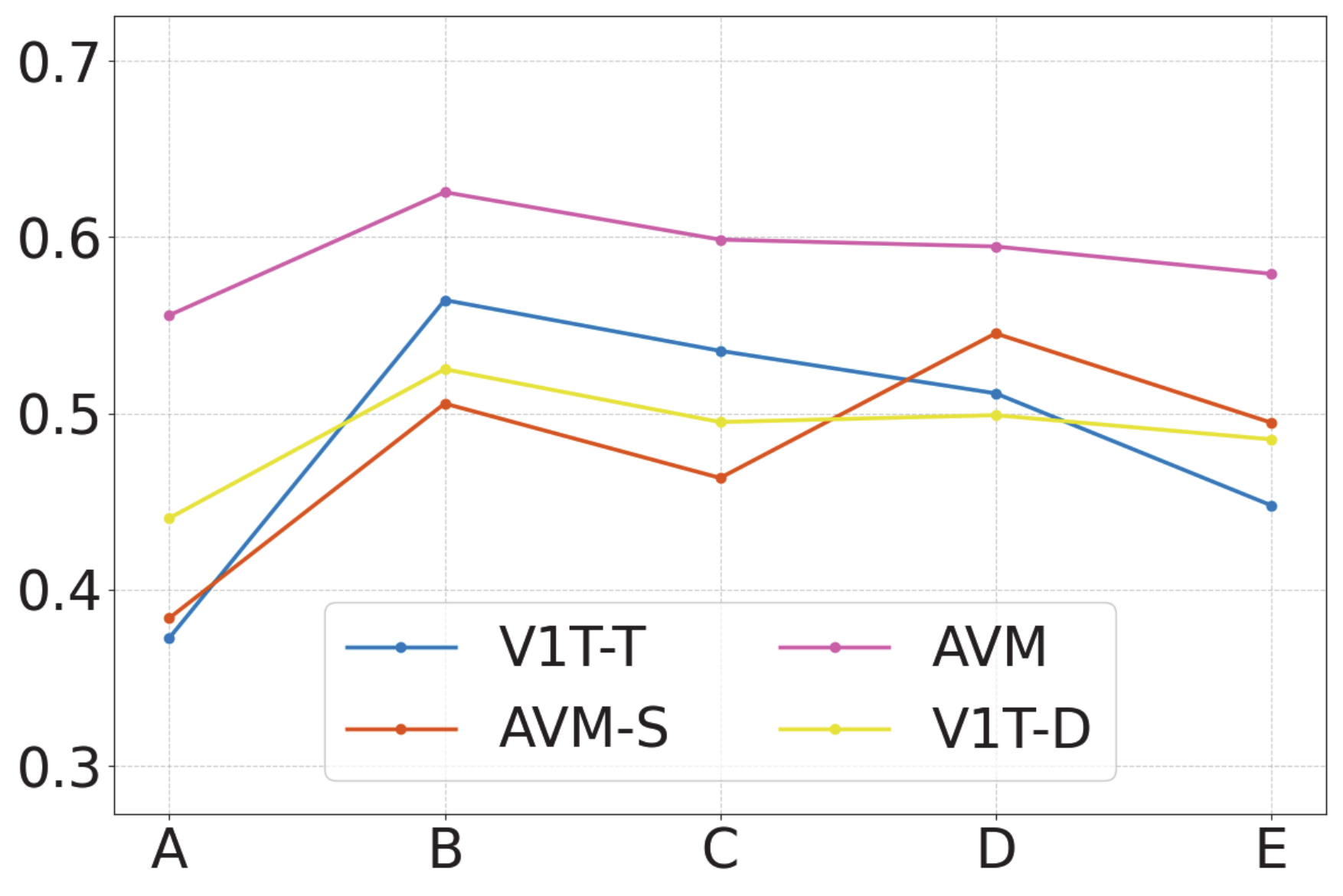}
    \vspace{0.5em}
    \includegraphics[width=\linewidth]{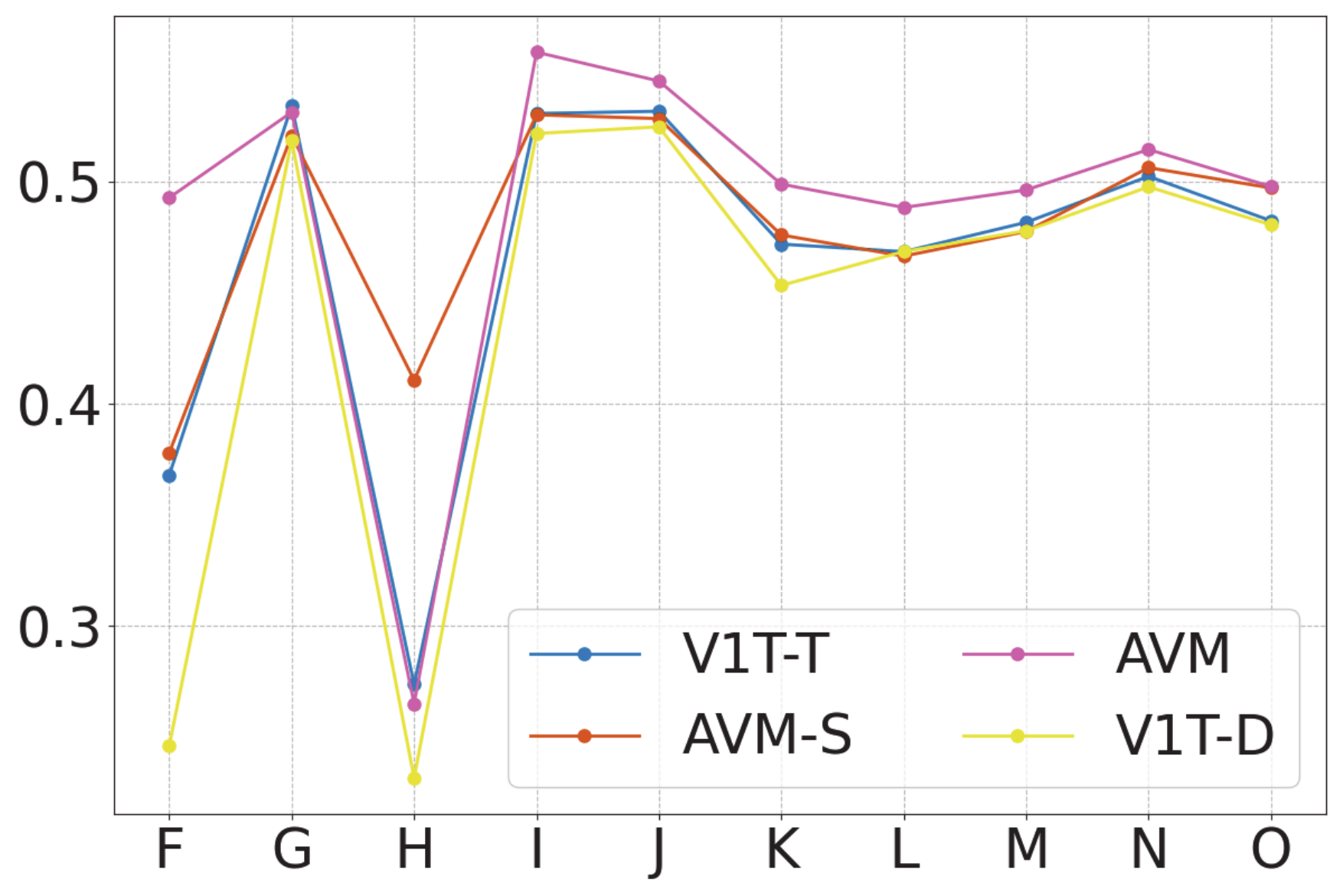}
    \caption*{(b) Average Correlation}
\end{subfigure}
\hfill
\begin{subfigure}[t]{0.32\textwidth}
    \centering
    \includegraphics[width=\linewidth]{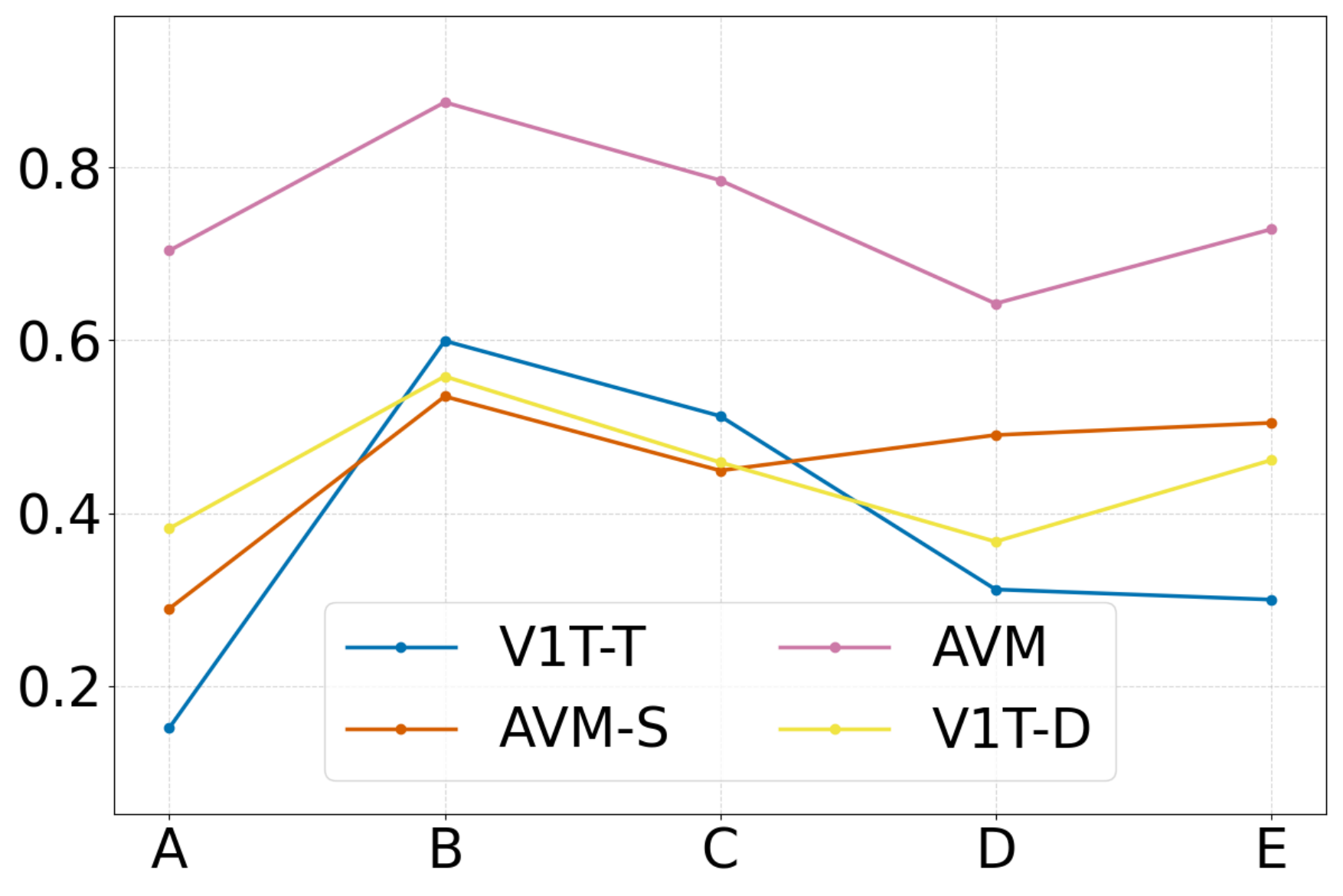}
    \vspace{0.5em}
    \includegraphics[width=\linewidth]{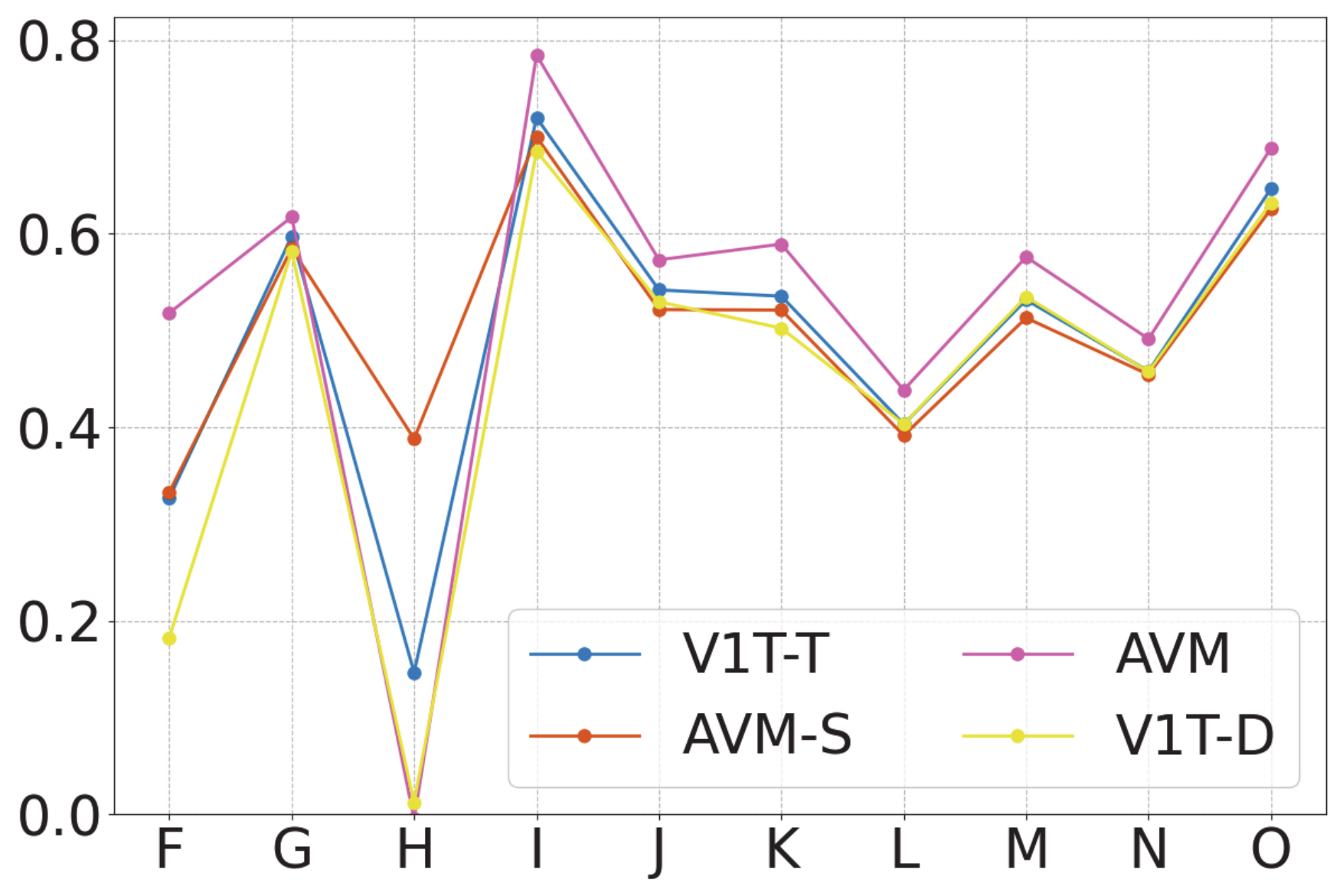}
    \caption*{(c) FEVE}
\end{subfigure}

\caption{The tuning ability of the AVM model under different input conditions. The top three figures show the results for dataset S, and the bottom three figures show the results for dataset F. From left to right, these figures show the single-trial correlation, average correlation, and explained variance, respectively. Each figure includes four structures: V1T-D, V1T-T, AVM-S, and AVM. The x-axis represents each mouse, and the y-axis represents the predicted value.}
\label{the new stimuli figure}
\end{figure*}

\textbf{AVM can capture the response shift caused by different input changes in the same individual.} To evaluate AVM's capability to capture condition-specific shifts in neural responses under input variation, we conducted experiments simulating visual state transitions within individual animals. Instead of retraining the entire system, AVM introduces condition-sensitive modulation paths embedded within each representational block. These paths locally reshape output activations in response to new stimuli while preserving a stable structural stream, supporting structure-function decoupling in cortical modeling.

As shown in Fig~\ref{F-single-mouse}, AVM consistently outperformed the baseline V1T across all mice in Dataset-F, achieving higher single-trial correlation ($\rho_{\text{trial}}$), trial-averaged correlation ($\rho_{\text{avg}}$), and fraction of explained variance (FEVE). These improvements demonstrate the model’s ability to fine-tune response predictions without altering the core representation pathway. Furthermore, when deploying a shared representation across mice in Dataset-F and Dataset-S (Fig~\ref{the new stimuli figure}), AVM maintained robust generalization performance and surpassed all comparative variants, including the pre-trained fine-tuning model (V1T-T), non-modulated baseline (V1T-D), and shared modulation version (AVM-S). Notably, AVM achieved a $>$30\% improvement in FEVE on Dataset-S, indicating that localized, condition-specific paths enable more expressive regulation of stimulus-driven variability.To ensure the stability of these results, we ran experiments using five different random seeds, and the standard deviations of the key metrics were consistently less than 0.001, demonstrating the robustness of AVM's performance.

\begin{figure}[h]
\begin{center}
\centerline{\includegraphics[width=0.85\columnwidth]{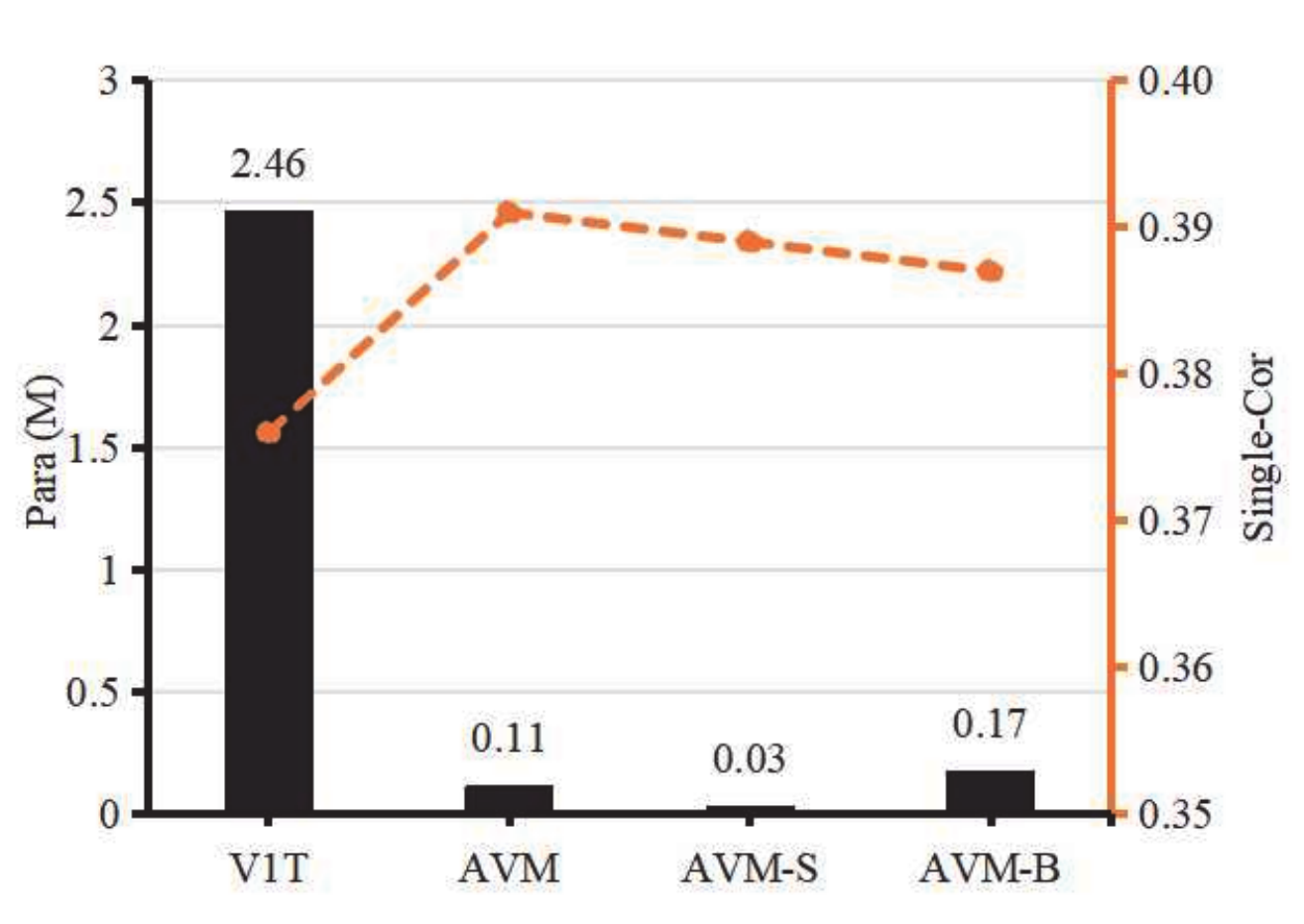}}
\caption{The number of trainable parameters. Comparison of the number of trainable parameters of our proposed AVM core and V1T core.}
\label{The number of trainable parameters}
\end{center}
\vskip -0.5in
\end{figure}

To assess architectural efficiency, we further compared the number of trainable parameters required by AVM variants and the V1T baseline (Fig ~\ref{The number of trainable parameters}). While V1T involves over 2.46M trainable parameters, AVM and its variants drastically reduce this cost to as low as 0.03M for AVM-S and 0.11M for AVM, with minimal compromise in predictive accuracy. 

\subsection{Adaptive Modeling Under Individual Variation}

\textbf{AVM enables condition-aware generalization across individuals with minimal structural adjustment.} To assess how well AVM models subject-specific response patterns while maintaining representational consistency, we design an individual generalization task in which neural responses from held-out subjects must be predicted based on knowledge learned from others. Specifically, for each subject in the dataset, we train the model using data from all remaining individuals and adapt it to the target subject using a small amount of subject-specific tuning.

This setting reflects a biologically plausible scenario where cortical encoding remains largely invariant, while individual-level variability is captured through lightweight, context-dependent adaptation. The evaluation is conducted on both Dataset-S and Dataset-F, and compared against three baselines: V1T-D (training from scratch), V1T-T (pretraining and full-model fine-tuning), and AVM-S (shared adaptation across subjects without conditioning).

As shown in Table~\ref{across animal-F-table} and Table~\ref{across animal-S-table}, AVM consistently outperforms all baselines across all three metrics—single-trial correlation, average correlation, and explained variance (FEVE). The gains are particularly pronounced on Dataset-F, with average improvements ranging from 1.5\% to 3.0\%. Moreover, AVM achieves this performance with minimal subject-specific parameterization, highlighting the efficiency and scalability of condition-aware response modulation. The comparison with AVM-S further demonstrates the importance of individual conditioning: adaptive tuning that accounts for subject identity outperforms shared adjustment strategies, underscoring the utility of contextual modulation in capturing biological variability.
\begin{table*}[t]
\begin{center}
\begin{tabular}{l|cccccccccc}
\toprule
Core & F & G & H & I & J & K & L & M & N & O \\
\midrule
\textbf{$\rho_{trial}$} \\
V1T-D    & 0.3153& 0.3815& 0.2743& 0.3859& 0.3856& 0.3713& 0.2895& 0.3688& 0.3510& 0.3682 \\
V1T-T    & 0.3189& 0.3790& 0.3628& 0.3817& 0.3895 & 0.3481& 0.2859& 0.3629& 0.3501& 0.3637\\
\textbf{AVM-S} & \textbf{0.3252} & \textbf{0.3862} & \textbf{0.3560} & \textbf{0.3883} & \textbf{0.3917} & \textbf{0.3666} & \textbf{0.2936} & \textbf{0.3703} & \textbf{0.3512} & \textbf{0.3738} \\
\textbf{AVM} & \textbf{0.3264} & \textbf{0.3860} & \textbf{0.3673} & \textbf{0.3898} & \textbf{0.3918} & \textbf{0.3694} & \textbf{0.2951} & \textbf{0.3691} & \textbf{0.3527} & \textbf{0.3748} \\
\midrule
\textbf{$\rho_{avg}$} \\
V1T-D    & 0.5141& 0.5605& 0.2303& 0.5650& 0.5718& 0.5251& 0.5120& 0.5254& 0.5398& 0.5120 \\
V1T-T    & 0.5307& 0.5692& 0.4752& 0.5828& 0.5828 & 0.5245& 0.5237& 0.5356& 0.5453& 0.5287\\
\textbf{AVM-S} & \textbf{0.5322} & \textbf{0.5795} & \textbf{0.4600} & \textbf{0.5803} & \textbf{0.5855} & \textbf{0.5361} & \textbf{0.5232} & \textbf{0.5398} & \textbf{0.5487} & \textbf{0.5306} \\
\textbf{AVM} & \textbf{0.5341} & \textbf{0.5794} & \textbf{0.4711} & \textbf{0.5841} & \textbf{0.5854} & \textbf{0.5345} & \textbf{0.5256} & \textbf{0.5382} & \textbf{0.5526} & \textbf{0.5317} \\
\midrule
\textbf{FEVE} \\
V1T-D    & 0.5305& 0.6685& -0.0037& 0.8202& 0.6198& 0.6827& 0.4829& 0.6648& 0.5423& 0.7413 \\
V1T-T    & 0.5615& 0.6851& 0.4798& 0.8372& 0.6371 & 0.6199& 0.4821& 0.6449& 0.5461& 0.7305\\
\textbf{AVM-S} & \textbf{0.5766} & \textbf{0.6992} & \textbf{0.4635} & \textbf{0.8464} & \textbf{0.6419} & \textbf{0.6696} & \textbf{0.4943} & \textbf{0.6736} & \textbf{0.5444} & \textbf{0.7627} \\
\textbf{AVM} & \textbf{0.5837} & \textbf{0.7201} & \textbf{0.4661} & \textbf{0.8608} & \textbf{0.6425} & \textbf{0.6811} & \textbf{0.4999} & \textbf{0.6649} & \textbf{0.5476} & \textbf{0.7692} \\
\bottomrule
\end{tabular}
\end{center}
\caption{Experimental results of regulatory ability across individuals in dataset F. Four models are compared in the table: V1T-D, V1T-T, AVM-S, and AVM. The three rows in the table correspond to three indicators, respectively.}
\label{across animal-F-table}
\end{table*}

\begin{table}[t]
\begin{center}
\fontsize{9}{11}\selectfont  
\begin{tabular}{l|ccccc}
\toprule
Core & A & B & C & D & E \\
\midrule
\textbf{$\rho_{trial}$} \\
V1T-D    & 0.3607& 0.4176& 0.3947& 0.4132& 0.3382 \\
V1T-T    & 0.3787& 0.4522& 0.4124& 0.4145& 0.3833\\
\textbf{AVM-S} & \textbf{0.3849} & \textbf{0.4575} & \textbf{0.4157} & \textbf{0.4258} & \textbf{0.3897} \\
\textbf{AVM} & \textbf{0.3855} & \textbf{0.4588} & \textbf{0.4199} & \textbf{0.4264} & \textbf{0.3916} \\
\midrule
\textbf{$\rho_{avg}$} \\
V1T-D    & 0.5343& 0.6072& 0.5812& 0.6061& 0.5091 \\
V1T-T    & 0.5669& 0.6519& 0.6181& 0.6117& 0.5896\\
\textbf{AVM-S} & \textbf{0.5744} & \textbf{0.6574} & \textbf{0.5185} & \textbf{0.6230} & \textbf{0.5969} \\
\textbf{AVM} & \textbf{0.5751} & \textbf{0.6585} & \textbf{0.6233} & \textbf{0.6230} & \textbf{0.5972} \\
\midrule
\textbf{FEVE} \\
V1T-D    & 0.6467& 0.7927& 0.7223& 0.6366& 0.5551 \\
V1T-T    & 0.7269& 0.9483& 0.8157& 0.6551& 0.7520\\
\textbf{AVM-S} & \textbf{0.7553} & \textbf{0.9711} & \textbf{0.8271} & \textbf{0.6936} & \textbf{0.7699} \\
\textbf{AVM} & \textbf{0.7500} & \textbf{0.9755} & \textbf{0.8424} & \textbf{0.6916} & \textbf{0.7752} \\
\bottomrule
\end{tabular}
\end{center}
\caption{Experimental results of regulatory ability across individuals in dataset S. }
\label{across animal-S-table}
\end{table}

\begin{table}[ht]
\begin{center}
\begin{tabular}{l|ccc}
\toprule
Core & $\rho_{trial}$ & $\rho_{avg}$ & \textbf{FEVE} \\
\midrule
DataSet F \\
\midrule
LN     & 0.2230 & --- & --- \\
Lurz    & 0.3090 & --- & --- \\
Deng    & --- & 0.600 & 0.558 \\
V1T-D    & 0.3607& 0.5388  & 0.6363 \\
V1T-T    & 0.3761& 0.5954 & 0.6662 \\
\textbf{AVM-S}    & \textbf{0.3893}& \textbf{0.6104}& \textbf{0.7515}\\
\textbf{AVM-B}    & \textbf{0.3873}& \textbf{0.6076}& \textbf{0.7467}\\
\textbf{AVM}    & \textbf{0.3906}& \textbf{0.6114}& \textbf{0.7536}\\
\bottomrule
\end{tabular}
\end{center}
\caption{Prediction performance under different environmental distribution conditions. This experiment compares four baseline models, namely LN linear model, CNN convolutional model, V1T-D and V1T-T. The experiments are conducted on dataset F. AVM, AVM-S and AVM-B represent three different frameworks proposed in this paper.}
\label{across dataset-F-table}
\end{table}

\subsection{Adaptive Modeling Under Environmental Shift}
\textbf{AVM enables robust neural response prediction across environmental distribution shifts.} To assess its generalization ability, we conduct a cross-dataset adaptation experiment. The model is trained on Dataset-S to obtain stable visual representations and then adapted to Dataset-F using lightweight contextual modulation. This simulates adapting to a new stimulus distribution while preserving core representational consistency. We compare AVM against several baselines, including a linear-nonlinear (LN) model, a convolutional predictor (Lurz), a method from Deng et al., and two Transformer architectures—V1T-D (trained from scratch) and V1T-T (fully fine-tuned). We also evaluate three AVM structural variants to assess different modulation mechanisms.

Notably, despite being trained solely on Dataset-S, AVM achieves strong performance on Dataset-F after lightweight adaptation—outperforming models trained directly on Dataset-F, such as Lurz and Deng. As shown in Table~\ref{across dataset-F-table}, AVM achieves 0.3906 in single-trial correlation, 0.6114 in average correlation, and 0.7536 in FEVE, with a 9.1\% improvement in explained variance over the V1T-T baseline. AVM improves single-trial correlation by over 8\% compared to Lurz, and FEVE by more than 19 percentage points compared to Deng. These results demonstrate that AVM achieves high generalization performance in new environments through structure-preserving learning and condition-aware modulation. Ablation studies (see appendix) further show that tuning modulation strength and bottleneck dimensions is crucial for robust adaptation. The AVM-B variant, however, suffers from mild oversharing and reduced layer specialization, leading to performance saturation and a decline in overall effectiveness.

\section{Conclusion}



This work introduces the Adaptive Visual Model (AVM), a structure-function decoupled framework for neural response modeling under variable biological and environmental conditions. By separating stable sensory encoding from flexible condition-driven modulation, AVM enables generalizable cortical response prediction without mixing representational learning and context adaptation.
We evaluate AVM across three generalization settings: stimulus variation, cross-individual transfer, and domain adaptation between environments using two large-scale mouse V1 datasets. In all scenarios, AVM outperforms strong baselines while maintaining efficiency and interpretability. These results show that condition-aware modulation atop a shared scaffold enables robust adaptation to input and subject variability without compromising structural stability.
This study highlights the importance of modeling functional modulation as an explicit computational goal in neuroscience-inspired architectures. Future work will extend this to higher cortical regions and explore real-time or closed-loop applications, advancing biologically grounded neural prediction.

\appendix

\section{Acknowledgments}
This work was supported in part by the National Natural Science Foundation of China (NSFC) under Grant (No. 62476035, 62206037, and U24B20140), and the Young Elite Scientists Sponsorship Program by CAST under Grant 2024QNRC001.


\bibliography{aaai2026}

\newpage
\appendix
\onecolumn
\section{Appendix}\label{app:myappendix}
\subsection{A.Effect of Modulation Configuration on Context-Aware Adaptation}
To evaluate the flexibility and robustness of AVM’s modulation mechanism, we conducted a series of ablation studies examining two key architectural factors: the scaling weight applied to modulation outputs (modulation strength) and the internal bottleneck dimension used in each modulation component (modulation capacity). All experiments were performed under a fixed encoding backbone to isolate the impact of contextual modulation structure. Results are reported on both Dataset-F and Dataset-S to assess consistency across environments.

\textbf{Modulation strength.}
We first varied the output scaling weight to assess how modulation strength affects prediction quality. As shown in Table~\ref{EX5F} and Table~\ref{EX5S}, performance peaks when the weight is set to 1.0 on Dataset-F and 2.0 on Dataset-S, indicating that optimal integration of contextual signals may vary slightly across environments. Under-modulation (e.g., weight = 0.1) leads to under-adjustment and limited responsiveness to contextual variation, while over-modulation (e.g., weight = 2.0 on Dataset-F) risks degrading the representational stability by injecting excessive context-specific influence.

\textbf{Modulation capacity.}
We also assessed the impact of the bottleneck dimension, which controls the internal capacity of each modulation layer. Results reveal a stable performance plateau across a broad range of settings (5–100), with peak values achieved at dimension = 31 on Dataset-F and dimension = 100 on Dataset-S. This demonstrates that AVM does not rely on large or overparameterized modulation modules to capture context-specific adjustments. Instead, it achieves strong generalization and adaptation through compact, structure-aware modulation design.

Together, these findings confirm that the condition-aware modulation in AVM is not a residual auxiliary mechanism but a functional and flexible component that enables scalable response adjustment. The performance sensitivity to modulation configuration further supports our core hypothesis: adaptive cortical modeling can be achieved by structurally disentangling stable encoding from dynamic contextual influences.


\begin{table}[h]
\vskip -0.1in
\begin{center}
\begin{scriptsize}  
\begin{tabular}{l|ccc}
\toprule
 & $\rho_{trial}$ & $\rho_{avg}$ &\textbf{FEVE} \\
\midrule
Weight \\
\midrule
0.1    & 0.3637$\pm$ 0.001& 0.5640$\pm$ 0.003& 0.5507$\pm$ 0.02 \\
0.5    & 0.3691$\pm$ 0.001& 0.5797$\pm$ 0.003& 0.5893$\pm$ 0.02 \\
2.0    & 0.3782$\pm$ 0.001& 0.5975$\pm$ 0.003& 0.6824$\pm$ 0.02 \\
\textbf{1.0}    & \textbf{0.3907$\pm$ 0.001}& \textbf{0.6114$\pm$ 0.002}& \textbf{0.7536$\pm$ 0.02}\\
\midrule
Dimension \\
\midrule
1    & 0.3861$\pm$ 0.001& 0.6053$\pm$ 0.003& 0.7219$\pm$ 0.03 \\
5    & 0.3873$\pm$ 0.001& 0.6074$\pm$ 0.003& 0.7316$\pm$ 0.03 \\
50    & 0.3877$\pm$ 0.001& 0.6079$\pm$ 0.003& 0.7339$\pm$ 0.03 \\
100    & 0.3892$\pm$ 0.001& 0.6099$\pm$ 0.003& 0.7504$\pm$ 0.03 \\
\textbf{31}    & \textbf{0.3907$\pm$ 0.001}& \textbf{0.6114$\pm$ 0.002}& \textbf{0.7536$\pm$ 0.02}\\
\bottomrule
\end{tabular}
\end{scriptsize}  
\end{center}
\vskip -0.15in
\caption{\textbf{Ablation study on modulation strength and bottleneck dimension.}
We evaluate the impact of modulation path configuration on AVM’s adaptive performance by varying the scaling weight (0.1, 0.5, 1.0, 2.0) and the bottleneck dimension (1, 5, 31, 50, 100). All experiments are conducted on Dataset F under a frozen main encoder. Results indicate that optimal performance is achieved with a balanced modulation strength (weight = 1.0) and a compact intermediate size (dimension = 31).}
\label{EX5F}
\end{table}

\begin{table}[h]
\vskip -0.1in
\begin{center}
\begin{scriptsize}  
\begin{tabular}{l|ccc}
\toprule
 & $\rho_{trial}$ & $\rho_{avg}$ &\textbf{FEVE}\\
\midrule
Dimension \\
\midrule
1    & 0.3455$\pm$ 0.001& 0.5099$\pm$ 0.003& 0.5589$\pm$ 0.006 \\
5    & 0.3528$\pm$ 0.001& 0.5196$\pm$ 0.001& 0.5764$\pm$ 0.006 \\
31    & 0.3637$\pm$ 0.001& 0.5352$\pm$ 0.001& 0.6054$\pm$ 0.006 \\
50    & 0.3659$\pm$ 0.001& 0.5377$\pm$ 0.001& 0.6121$\pm$ 0.006 \\
\textbf{100}    & \textbf{0.3689$\pm$ 0.001}& \textbf{0.5407$\pm$ 0.001}& \textbf{0.6225$\pm$ 0.006}\\
\midrule
Weight \\
\midrule
0.1    & 0.3562$\pm$ 0.001& 0.5245$\pm$ 0.001& 0.5854$\pm$ 0.006 \\
0.5    & 0.3631$\pm$ 0.001& 0.5335$\pm$ 0.001& 0.6052$\pm$ 0.006 \\
1.0    & 0.3637$\pm$ 0.001& 0.5352$\pm$ 0.001& 0.6054$\pm$ 0.006 \\
\textbf{2.0}    & \textbf{0.3647$\pm$ 0.001}& \textbf{0.5354$\pm$ 0.001}& \textbf{0.6094$\pm$ 0.006}\\
\bottomrule
\end{tabular}
\end{scriptsize}  
\end{center}
\vskip -0.15in
\caption{\textbf{Experiments by modifying weights and dimensions.}
We perform ablation experiments by modifying the weights and dimensions of the new structures generated by environmental stimuli. These experiments are conducted on dataset S. For each dataset, we test four different weight configurations: 0.1, 0.5, 1.0, and 2.0, and five different dimensions: 1, 5, 31, 50, and 100. }
\label{EX5S}
\end{table}

\begin{table}[H]
\begin{center}
\begin{tabular}{c|cccccccccc}
\toprule
Weight & F & G & H & I & J & K & L & M & N & O \\
\midrule
\textbf{$\rho_{trial}$} \\
0.1 & 0.3006 & 0.4015 & 0.3290 & 0.3885 & 0.4052 & 0.3642 & 0.2996 & 0.3901 & 0.3823 & 0.3758 \\
0.5 & 0.3130 & 0.4007 & 0.3343 & 0.3984 & 0.4048 & 0.3759 & 0.3084 & 0.3865 & 0.3836 & 0.3851 \\
1.0 & 0.3314 & 0.4274 & 0.3614 & 0.4165 & 0.4290 & 0.3957 & 0.3233 & 0.4158 & 0.4066 & 0.3997 \\
2.0 & 0.3208 & 0.4172 & 0.3338 & 0.4046 & 0.4182 & 0.3816 & 0.3125 & 0.4063 & 0.3958 & 0.3911 \\
\midrule
\textbf{$\rho_{avg}$} \\
0.1 & 0.5145 & 0.6075 & 0.4109 & 0.6091 & 0.6164 & 0.5702 & 0.5568 & 0.5934 & 0.6032 & 0.5570 \\
0.5 & 0.5385 & 0.6186 & 0.4299 & 0.6327 & 0.6255 & 0.5939 & 0.5776 & 0.5917 & 0.6126 & 0.5764 \\
1.0 & 0.5661 & 0.6536 & 0.4759 & 0.6567 & 0.6603 & 0.6192 & 0.5994 & 0.6356 & 0.6484 & 0.5987 \\
2.0 & 0.5529 & 0.6397 & 0.4434 & 0.6415 & 0.6470 & 0.6047 & 0.5842 & 0.6290 & 0.6412 & 0.5912 \\
\midrule
\textbf{FEVE} \\
0.1 & 0.5090 & 0.7720 & 0.2049 & 0.8925 & 0.6922 & 0.7001 & 0.5235 & 0.7772 & 0.6157 & 0.7947 \\
0.5 & 0.4856 & 0.6855 & 0.2367 & 0.8181 & 0.5981 & 0.6932 & 0.4200 & 0.6824 & 0.5451 & 0.7283 \\
1.0 & 0.6179 & 0.8611 & 0.4338 & 0.9898 & 0.7729 & 0.8172 & 0.6014 & 0.8611 & 0.7059 & 0.8752 \\
2.0 & 0.5517 & 0.7992 & 0.3023 & 0.9113 & 0.7017 & 0.7439 & 0.5477 & 0.7857 & 0.6460 & 0.8347 \\
\bottomrule
\end{tabular}
\end{center}
\caption{\textbf{Impact of Modulation Weight on Adaptive Performance (Dataset-F).}
We report trial-wise correlation, average correlation, and explained variance (FEVE) for each mouse in Dataset-F under four modulation weight settings (0.1, 0.5, 1.0, 2.0). Results highlight that moderate scaling (1.0) achieves the most consistent performance across individuals.}
\label{weight_ablation-F}
\end{table}

\begin{table}[H]
\begin{center}
\begin{tabular}{c|cccccccccc}
\toprule
Weight & F & G & H & I & J & K & L & M & N & O \\
\midrule
\textbf{$\rho_{trial}$} \\
1 & 0.3252 & 0.4248 & 0.3481 & 0.4118 & 0.4266 & 0.3893 & 0.3203 & 0.4130 & 0.4045 & 0.3974 \\
5 & 0.3264 & 0.4261 & 0.3446 & 0.4155 & 0.4284 & 0.3916 & 0.3219 & 0.4146 & 0.4050 & 0.3993 \\
31 & 0.3314 & 0.4274 & 0.3614 & 0.4165 & 0.4290 & 0.3957 & 0.3233 & 0.4158 & 0.4066 & 0.3997 \\
50 & 0.3295 & 0.4273 & 0.3432 & 0.4149 & 0.4287 & 0.3920 & 0.3225 & 0.4153 & 0.4047 & 0.3987 \\
100 & 0.3312 & 0.4278 & 0.3458 & 0.4170 & 0.4298 & 0.3946 & 0.3223 & 0.4159 & 0.4066 & 0.4010 \\
\midrule
\textbf{$\rho_{avg}$} \\
1 & 0.5595 & 0.6507 & 0.4448 & 0.6522 & 0.6577 & 0.6140 & 0.5964 & 0.6336 & 0.6460 & 0.5978 \\
5 & 0.5590 & 0.6525 & 0.4529 & 0.6549 & 0.6599 & 0.6150 & 0.5988 & 0.6346 & 0.6470 & 0.5991 \\
31 & 0.5661 & 0.6536 & 0.4759 & 0.6567 & 0.6603 & 0.6192 & 0.5994 & 0.6356 & 0.6484 & 0.5987 \\
50 & 0.5613 & 0.6540 & 0.4511 & 0.6571 & 0.6606 & 0.6153 & 0.5996 & 0.6353 & 0.6462 & 0.5990 \\
100 & 0.5653 & 0.6531 & 0.4632 & 0.6573 & 0.6609 & 0.6181 & 0.5994 & 0.6359 & 0.6468 & 0.5994 \\
\midrule
\textbf{FEVE} \\
1 & 0.5942 & 0.8522 & 0.2321 & 0.9714 & 0.7653 & 0.7938 & 0.5902 & 0.8521 & 0.6975 & 0.8702 \\
5 & 0.5993 & 0.8573 & 0.2741 & 0.9850 & 0.7729 & 0.7985 & 0.5967 & 0.8574 & 0.7007 & 0.8740 \\
31 & 0.6179 & 0.8611 & 0.4338 & 0.9898 & 0.7729 & 0.8172 & 0.6014 & 0.8611 & 0.7059 & 0.8752 \\
50 & 0.6108 & 0.8621 & 0.2814 & 0.9831 & 0.7724 & 0.8033 & 0.5985 & 0.8599 & 0.6974 & 0.8704 \\
100 & 0.6193 & 0.8631 & 0.3924 & 0.9933 & 0.7759 & 0.8144 & 0.5970 & 0.8644 & 0.7042 & 0.8804 \\
\bottomrule
\end{tabular}
\end{center}
\caption{\textbf{Impact of Bottleneck Dimension on Adaptive Performance (Dataset-F).}
We evaluate the effect of bottleneck dimension size (1, 5, 31, 50, 100) on AVM's predictive performance across mice in Dataset-F. Results suggest that a moderately compact representation (dimension = 31) yields optimal trade-offs between efficiency and adaptation capacity.
}
\label{dim_ablation-F}
\end{table}

\newpage
\begin{table}[H]
\begin{center}
\begin{tabular}{c|ccccc}
\toprule
Weight & A & B & C & D & E \\
\midrule
\textbf{$\rho_{trial}$} \\
0.1 & 0.3345 & 0.3937 & 0.3605 & 0.3542 & 0.3381 \\
0.5 & 0.3404 & 0.4011 & 0.3665 & 0.3641 & 0.3435 \\
1.0 & 0.3411 & 0.4013 & 0.3681 & 0.3648 & 0.3434 \\
2.0 & 0.3428 & 0.4014 & 0.3683 & 0.3661 & 0.3450 \\
\midrule
\textbf{$\rho_{avg}$} \\
0.1 & 0.4894 & 0.5627 & 0.5262 & 0.5285 & 0.5158 \\
0.5 & 0.4979 & 0.5722 & 0.5327 & 0.5408 & 0.5237 \\
1.0 & 0.4988 & 0.5741 & 0.5355 & 0.5429 & 0.5247 \\
2.0 & 0.5006 & 0.5725 & 0.5344 & 0.5437 & 0.5256 \\
\midrule
\textbf{FEVE} \\
0.1 & 0.5544 & 0.7182 & 0.6094 & 0.4689 & 0.5759 \\
0.5 & 0.5723 & 0.7407 & 0.6292 & 0.4926 & 0.5910 \\
1.0 & 0.5748 & 0.7394 & 0.6334 & 0.4901 & 0.5891 \\
2.0 & 0.5815 & 0.7403 & 0.6356 & 0.4960 & 0.5938 \\
\bottomrule
\end{tabular}
\end{center}
\caption{\textbf{Impact of Modulation Weight on Adaptive Performance (Dataset-S).}
This table presents the prediction performance on Dataset-S across five mice under varying modulation weights. A steady improvement is observed with increasing weight, peaking at 2.0.
}
\label{weight_ablation_S}
\end{table}

\begin{table}[H]
\begin{center}
\begin{tabular}{c|ccccc}
\toprule
Dimension & A & B & C & D & E \\
\midrule
\textbf{$\rho_{trial}$} \\
1 & 0.3268 & 0.3822 & 0.3487 & 0.3408 & 0.3292 \\
5 & 0.3324 & 0.3888 & 0.3563 & 0.3508 & 0.3357 \\
31 & 0.3411 & 0.4013 & 0.3681 & 0.3648 & 0.3434 \\
50 & 0.3424 & 0.4040 & 0.3692 & 0.3680 & 0.3462 \\
100 & 0.3452 & 0.4067 & 0.3719 & 0.3718 & 0.3492 \\
\midrule
\textbf{$\rho_{avg}$} \\
1 & 0.4787 & 0.5476 & 0.5113 & 0.5095 & 0.5025 \\
5 & 0.4863 & 0.5551 & 0.5210 & 0.5235 & 0.5120 \\
31 & 0.4988 & 0.5741 & 0.5355 & 0.5429 & 0.5247 \\
50 & 0.5002 & 0.5778 & 0.5372 & 0.5457 & 0.5276 \\
100 & 0.5029 & 0.5793 & 0.5400 & 0.5497 & 0.5315 \\
\midrule
\textbf{FEVE} \\
1 & 0.5396 & 0.6844 & 0.5764 & 0.4406 & 0.5535 \\
5 & 0.5527 & 0.7007 & 0.5962 & 0.4612 & 0.5714 \\
31 & 0.5748 & 0.7394 & 0.6334 & 0.4901 & 0.5891 \\
50 & 0.5765 & 0.7490 & 0.6358 & 0.5011 & 0.5979 \\
100 & 0.5871 & 0.7597 & 0.6487 & 0.5108 & 0.6062 \\
\bottomrule
\end{tabular}
\end{center}
\caption{\textbf{Impact of Bottleneck Dimension on Adaptive Performance (Dataset-S).}
We test bottleneck dimensions from 1 to 100 on Dataset-S. Performance metrics improve with larger dimensions, with the best results observed at the largest setting (dimension = 100).
}
\label{dim_ablation-S}
\end{table}

\newpage
\subsection{B. Experimental Results for Single-Trial Correlation on Mouse F}
\begin{table}[ht]
\centering
\begin{tabular}{lccc}
\hline
\textbf{Metric} & \textbf{Mean} & \textbf{Standard Deviation} & \textbf{95\% Confidence Interval} \\ \hline
\textbf{$\rho_{trial}$} & 0.3027 & $\pm$0.0005 & [0.3021, 0.3035] \\
\textbf{$\rho_{avg}$} & 0.4837 & $\pm$0.0004 & [0.4830, 0.4843] \\
\textbf{FEVE} & 0.5019 & $\pm$0.0009 & [0.5010, 0.5037] \\
\hline
\end{tabular}
\caption{Results of 5 Independent Random Experiments for Single-Trial Correlation on Mouse F}
\label{experimental_results}
\end{table}

Table \ref{experimental_results} presents the results of 5 independent random experiments conducted on Mouse F to evaluate the single-trial correlation, average correlation, and FEVE. The table reports the mean values, standard deviations, and 95\% confidence intervals for each metric. Specifically, the single-trial correlation ($\rho_{trial}$) has a mean of 0.3027 with a standard deviation of $\pm$0.0005, and the 95\% confidence interval is [0.3021, 0.3035]. For the average correlation ($\rho_{avg}$), the mean is 0.4837 with a standard deviation of $\pm$0.0004, and the confidence interval is [0.4830, 0.4843]. Finally, for FEVE, the mean is 0.5019, with a standard deviation of $\pm$0.0009, and the confidence interval is [0.5010, 0.5037]. These results indicate high consistency and stability across the repeated experiments.

\subsection{C. Parameter Description}
\begin{table}[ht]
\centering
\begin{tabular}{l l}
\hline
\textbf{Category}          & \textbf{Details}                                                   \\ \hline
\textbf{Model Hyperparameters}  & \textbf{}                                                          \\ 
Learning Rate              & 0.0016                                                             \\ 
Batch Size                 & 16                                                                  \\ 
Optimizer                  & AdamW                                                               \\ 
Loss Function              & Poisson Loss                                                        \\ 
Weight Decay               & 1e-4                                                                \\ 
Epochs                     & 400                                                                 \\ 
Early Stopping Patience    & 10 epochs                                                            \\ 
Learning Rate Decay        & 0.3                                                                  \\ 
Modulation Strength        & 1.0                                                                 \\ 
Bottleneck Dimension       & 31                                                                   \\ \hline
\textbf{Hardware Specifications} & \textbf{}                                                        \\ 
GPU                        & NVIDIA A100 40GB                                                     \\ 
RAM                        & 128 GB                                                              \\ 
Operating System           & Ubuntu 20.04.3 LTS                                                   \\ 
CUDA Version               & 11.8                                                                \\ 
Framework                  & PyTorch 2.4.0                                                       \\ \hline
\end{tabular}
\caption{Hyperparameters and Hardware Specifications}
\label{tab:hyperparams-hardware}
\end{table}
The table\ref{tab:hyperparams-hardware} displays the hyperparameters and hardware configuration used in this study. The hyperparameters section includes key settings for model training, such as the learning rate (0.0016), batch size (16), optimizer (AdamW), loss function (Poisson Loss), and weight decay (1e-4). The training process was set to 400 epochs with an early stopping strategy (10 epochs). Furthermore, the learning rate decay coefficient was set to 0.3, the modulation intensity to 1.0, and the bottleneck dimension to 31.

The hardware section lists the hardware configuration used for the experiments, employing an NVIDIA A100 40GB GPU and 128GB of RAM, running Ubuntu 20.04.3 LTS with CUDA version 11.8, and using PyTorch 2.4.0. This configuration ensured efficient experiment execution and supported the processing of large-scale datasets and model training.

\newpage
\subsection{D. CAMU Submodule Visualizations}
\begin{figure*}[htp]
	
	\begin{minipage}{0.32\linewidth}
		\vspace{3pt}
		\centerline{\includegraphics[width=\textwidth]{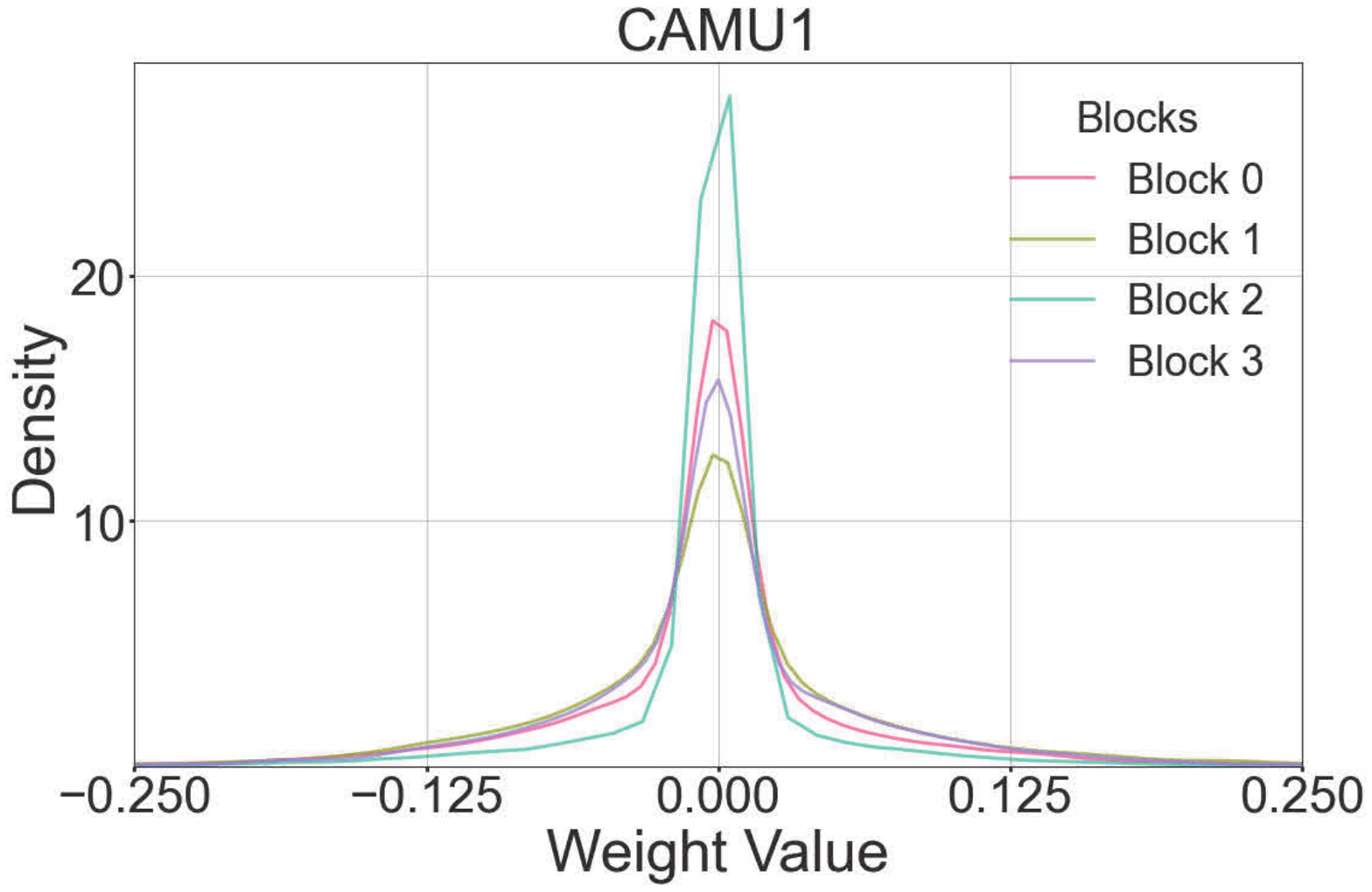}}
	\end{minipage}
	\begin{minipage}{0.32\linewidth}
		\vspace{3pt}
		\centerline{\includegraphics[width=\textwidth]{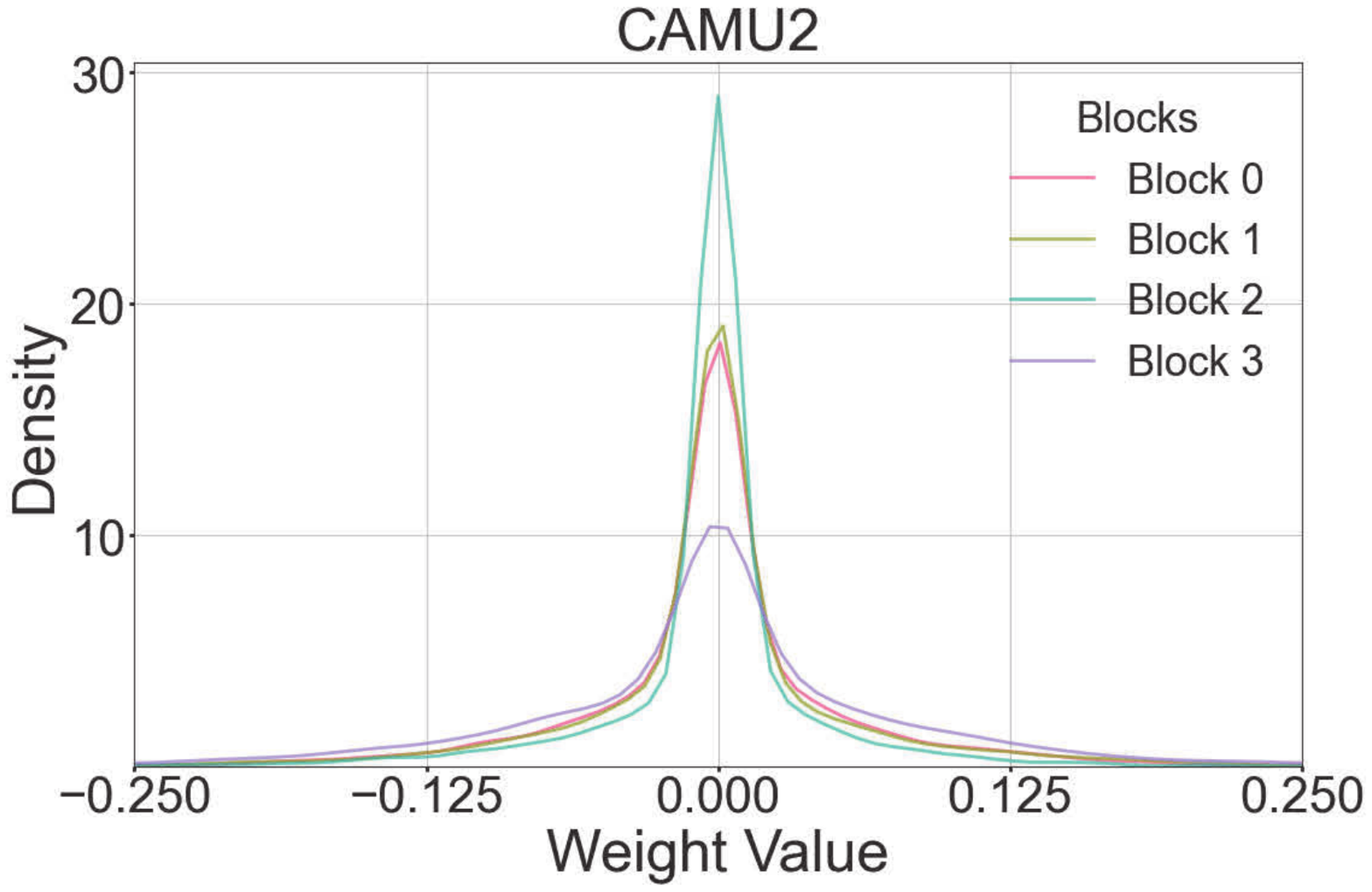}}
	\end{minipage}
	\begin{minipage}{0.32\linewidth}
		\vspace{3pt}
		\centerline{\includegraphics[width=\textwidth]{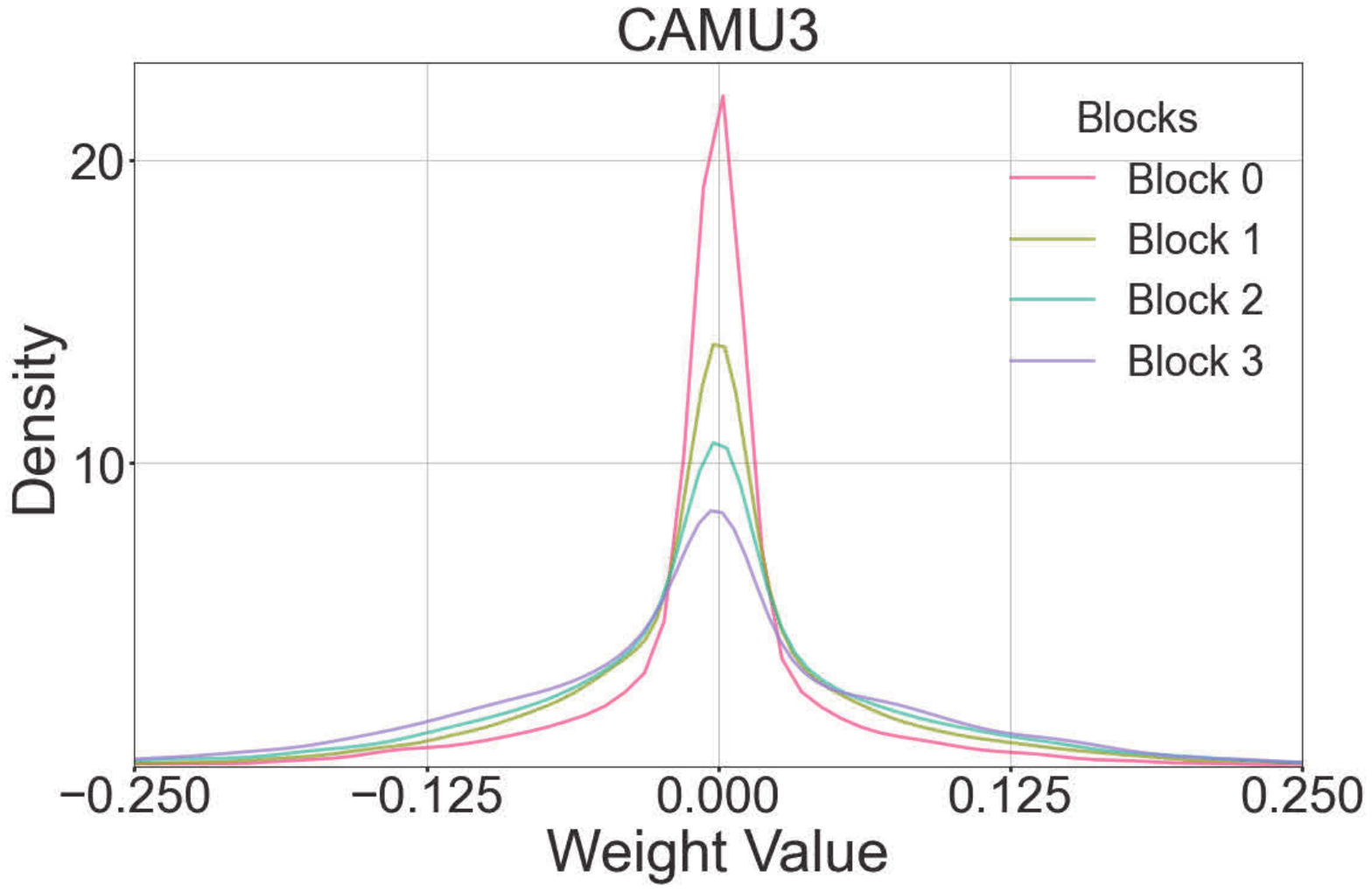}}
	\end{minipage}
 
	\caption{Visualization of the CAMU submodules (CAMU1, CAMU2, and CAMU3): These submodules represent different stages of modulation within the AVM framework. Each plot shows the weight distribution of the respective CAMU block, with varying levels of modulation strength across different blocks (Block 0 to Block 3). CAMU1, CAMU2, and CAMU3 highlight the diverse responses and weight adjustments as the model adapts to different stimuli and conditions.}

	\label{fig:camu_submodules}
\end{figure*}

Figure \ref{fig:camu_submodules} visualizes the three CAMU submodules in the AVM framework: CAMU1, CAMU2, and CAMU3. These submodules represent different modulation stages in the model, with each subplot showing the weight distribution of its respective CAMU block, reflecting different modulation intensities (from Block 0 to Block 3). CAMU1, CAMU2, and CAMU3 demonstrate how the model adjusts weights and responses across blocks to adapt to different stimuli and conditions. These visualizations provide a clearer view of how the model flexibly adapts while maintaining structural stability.

\end{document}